\documentclass{llncs}
\usepackage{amsmath, graphicx, booktabs, tabularx}
\usepackage{hyperref}

\graphicspath{{./figures/}}
\begin{document}

\title{Building Disease Detection Algorithms with Very Small Numbers of Positive Samples
\thanks{This paper was accepted by the International Conference on Medical Image Computing and Computer-Assisted Intervention -- MICCAI 2017. The final publication is available at Springer via \url{https://doi.org/10.1007/978-3-319-66179-7_54}.}}
\author{Ken C. L. Wong, Alexandros Karargyris, Tanveer Syeda-Mahmood,\\and Mehdi Moradi}
\institute{IBM Research -- Almaden Research Center, San Jose, CA, USA\\
\email{mmoradi@us.ibm.com}
}

\maketitle              

\begin{abstract}
Although deep learning can provide promising results in medical image analysis, the lack of very large annotated datasets confines its full potential. Furthermore, limited positive samples also create unbalanced datasets which limit the true positive rates of trained models. As unbalanced datasets are mostly unavoidable, it is greatly beneficial if we can extract useful knowledge from negative samples to improve classification accuracy on limited positive samples. To this end, we propose a new strategy for building medical image analysis pipelines that target disease detection. We train a discriminative segmentation model only on normal images to provide a source of knowledge to be transferred to a disease detection classifier. We show that using the feature maps of a trained segmentation network, deviations from normal anatomy can be learned by a two-class classification network on an extremely unbalanced training dataset with as little as one positive for 17 negative samples. We demonstrate that even though the segmentation network is only trained on normal cardiac computed tomography images, the resulting feature maps can be used to detect pericardial effusion and cardiac septal defects with two-class convolutional classification networks.
\end{abstract}
\section{Introduction}


Big data methods are raising the benchmarks of classification and detection challenges in computer vision. In medical imaging, however, given the huge variety of possible clinical conditions in an imaging modality, such as chest computed tomography (CT), it is extremely challenging to build a sufficiently large dataset with samples of abnormalities. As a result, most learning-based medical image analysis solutions focus on a narrow range of diseases. Apart from lack of generality, limited positive samples also create unbalanced datasets. If such datasets are directly used to train classifiers, low true positive rates can be expected.

Given these limits, we propose a new strategy for building medical image analysis pipelines that target disease detection. In simple terms, we try to extract useful knowledge from the negative samples and make use of such knowledge to improve classification on limited positive samples. This basic idea has been visited in computer vision with one-shot learning technics. In \cite{Journal:Fei:PAMI2006}, it was shown that generative models have the potential for one-shot learning, which allows learning of new categories with only a few samples when prior models of unrelated categories are given. Nevertheless, simple generative models may not be able to characterize delicate variations that separate normal from disease in medical imaging. Highly complex generative models potentially have the ability to model fine variations, but require very large training sets \cite{Conference:Goodfellow:NIPS2014}.

Our solution here is to use a discriminative model, which is trained on a segmentation task on normal images, as a source of knowledge to be transferred to the disease detection classifier. The use of a discriminative deep learner as the source of features has been explored in the past \cite{Journal:Shin:TMI2016,Conference:Moradi:ISBI2016}. In these previous works, the networks used for the feature sources are classifiers such as AlexNet and VGGNet which provide one label per image. In this work, we use a fully convolutional segmentation network to characterize normal (negative) anatomy of the heart without seeing any abnormal (positive) samples. Different from classification networks, segmentation networks output a label per pixel and thus provide the shapes and locations of the structures of interest. Therefore, although a trained segmentation network may not provide a prior as good as a generative model for one-shot learning, it has the potential to produce useful features for better classification accuracy on unbalanced data. We show that deviations from normal anatomy can be learned from the feature maps produced by this segmentation network using a two-class classification network, trained on an extremely unbalanced training dataset with as little as one positive for 17 negative samples.

The power of the method is partially due to the ability of the segmentation network to transfer image information to semantic pixel level labels provided by radiologists. As it was shown in \cite{Conference:Socher:NIPS2013}, cross-modal knowledge transfer from image to semantic text space can be used to detect unseen objects in new samples. The segmentation network used here is optimized from the work in \cite{Conference:Ronneberger:MICCAI2015} for segmenting normal heart anatomy from four chamber cardiac CT slices. By concatenating the low and high level features from the segmentation network, a classification network can be trained on unbalanced data.

To the best of our knowledge, this paper reports the first attempt at one-shot learning in medical imaging. We show that even though the segmentation network is trained only on normal CT images, the resulting feature maps can be used to detect pericardial effusion and cardiac septal defects with two-class convolutional classification networks.

\section{Methodology}

\begin{figure}[t]
    \centering
    \begin{minipage}[b]{1\linewidth}
      \centering
      \centerline{\includegraphics[width=1\linewidth]{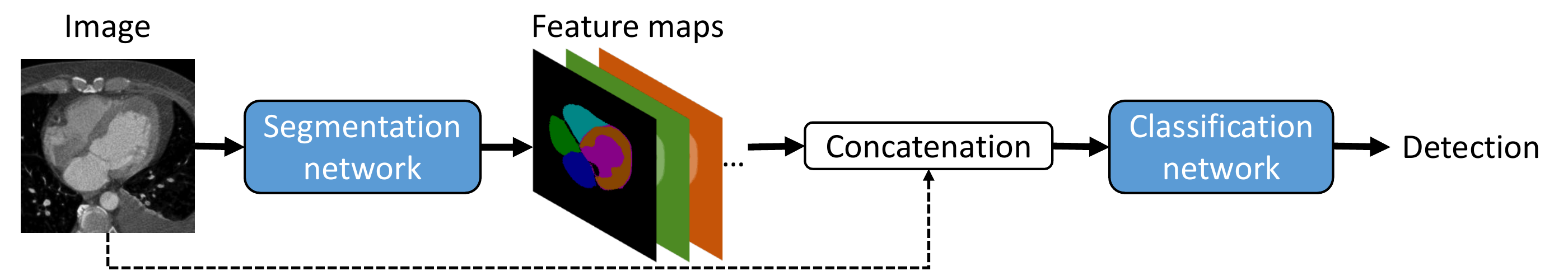}}
    \end{minipage}
    \caption{The overall disease detection framework combining a segmentation network and a classification network.}
    \label{fig:framework}
\end{figure}

\begin{figure}[t]
    \centering
    \begin{minipage}[b]{1\linewidth}
      \centering
      \centerline{\includegraphics[width=1\linewidth]{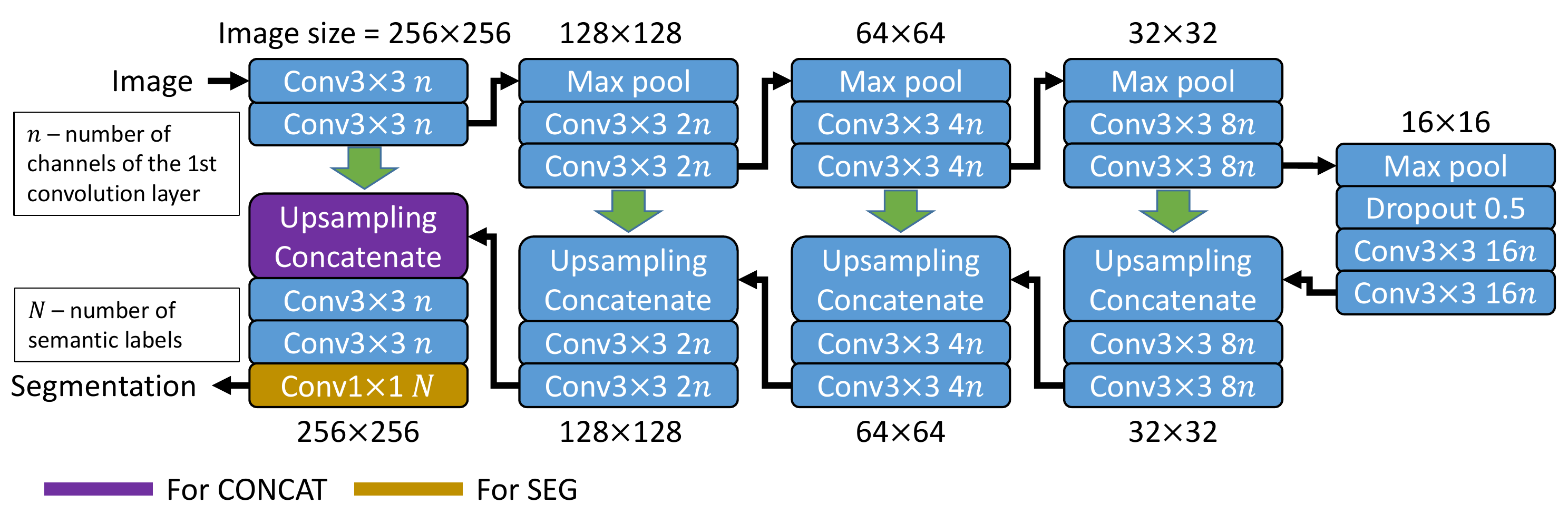}}
    \end{minipage}
    \caption{Architecture of the segmentation network. $n$ is the number of feature channels in the first convolution layer, and $N$ is the number of semantic labels.The green arrows indicate copying of feature maps. The feature maps from the last concatenation layer and the last convolution layer are input as different feature combinations.}
    \label{fig:segnet}
\end{figure}

The overall framework of our proposed model is shown in Fig. \ref{fig:framework}. The segmentation network is a modified version of the fully convolutional architecture in \cite{Conference:Ronneberger:MICCAI2015}. This architecture is used because of its capability for accurate segmentation with only a few training samples of biomedical images (e.g. 30 images in \cite{Conference:Ronneberger:MICCAI2015}). While there are many ways to utilize the knowledge of a trained segmentation network, to facilitate flexible exploration of feature combinations for classification, we use a two-class classification network modified based on the VGGNet architecture in \cite{Journal:Simonyan:arXiv2014}. We have a separate classifier per disease type.

\subsection{Network Architecture and Training}
\label{sec:architecture}

\textbf{Architecture of segmentation network: }The architecture of the segmentation network is depicted in Fig. \ref{fig:segnet}. The contracting path consists of repeated applications of two 3$\times$3 convolutions associated with exponential linear units (ELU), and a 2$\times$2 max pooling. The number of channels is doubled after each pooling. The expanding path consists of 2$\times$2 upsampling by repetition, a concatenation with the corresponding feature maps from the contracting path, and two ELU-associated 3$\times$3 convolutions that halve the number of channels. Padding is used for each convolution to ensure segmentation image of the same size. A final layer of 1$\times$1 convolution is used to map the feature maps to the number of interested semantic labels. The same as \cite{Conference:Ronneberger:MICCAI2015}, we have four max pooling and four upsampling layers. Instead of using 64 channels ($n=64$) for the first convolution layer, 16 channels ($n=16$) is used as it provides high accuracy with fewer parameters (Section \ref{sec:exp:seg}).

\textbf{Architecture of classification network: }For the classification network, an architecture with 16 weight layers is used, which consists of 13 3$\times$3 convolution layers followed by three fully-connected layers. A max pooling layer (five in total) is placed for every two or three convolution layers. This is similar to VGGNet in \cite{Journal:Simonyan:arXiv2014} but modified for two-class classification.

\textbf{Training strategy: }For the segmentation network, pixel-wise softmax is applied to the output of the last convolution layer to compute the probability maps for the semantic labels. Weighted cross entropy is then used to compute the loss function as:
\begin{align}
\label{eq:loss}
    L = - \sum_\mathbf{x} w_{l(\mathbf{x})} \log(p_{l(\mathbf{x})})
\end{align}
where $\mathbf{x}$ is the pixel position and $l(\mathbf{x})$ is the corresponding ground truth label. $p_{l(\mathbf{x})}$ is the softmax probability of the channel corresponds to $l(\mathbf{x})$. $w_{l(\mathbf{x})}$ is the weight computed from the pixel label frequencies of all training atlases as $w_l = 1 - f_l / \sum_k f_k$, with $f_k$ the number of pixels of ground truth label $k$. Although the weights are not summed to one, they reduce the influences of more frequently seen labels. The stochastic optimization algorithm Adam is used for fast convergence \cite{Journal:Kingma:arXiv2014}. The same training process is applied to the classification network, with the summation in (\ref{eq:loss}) removed and $l\in\{0,1\}$. Furthermore, $w_l = 1 - \left({f_l} / \sum_k {f_k}\right)^{1/4}$ to avoid over penalty of the negative samples as the ratio of the negative to the positive samples can be large.

For the combined architecture, the segmentation network is trained by the data which are not used by the training and testing of the classification network.

\subsection{Feature Combinations for Classification Network}
\label{sec:combinations}
Using the trained segmentation network, one obtains a variety of feature maps which can be used along with the original image for the classification step. To understand the benefits of using these features compared with using the original image alone for disease/normal classification, we tested a few arrangements:

\begin{itemize}
  \item \textbf{IMG}: Input only the original image to the classification network.
  \item \textbf{SEG}: Input the feature maps of the final 1$\times$1 convolution layer ($N$ channels) of the segmentation network to the classification network (Fig. \ref{fig:segnet}). This provides the high level segmentation feature maps.
  \item \textbf{IMG+SEG}: Same as SEG but also use the original image.
  \item \textbf{CONCAT}: Input the feature maps of the final upsampling and concatenation layer to the classification network (Fig. \ref{fig:segnet}). This provides feature maps ($3n$ channels) with both low and high level features.
  \item \textbf{IMG+CONCAT}: Same as CONCAT but also use the original image.
\end{itemize}

\subsection{Data}

We studied the performance of the proposed framework on disease detection with cardiac CT data. Our images include cases of pericardial effusion and septal defects which were diagnosed by radiologists from axial slices depicting the four chamber view of the heart. To avoid the computational costs of 3D networks, we performed analysis on 2D slices extracted from the CT data. One of the automatic methods for extracting the relevant slice is described in \cite{Conference:Moradi:ISBI2016}, which provides accuracy larger than 90\% in detecting the four chamber view.

Pericardial effusion is the abnormal accumulation of fluid in the pericardium which can impede cardiac filling \cite{Journal:Leary:BJR2010}. Septal defects are congenital abnormalities caused by malformations of the heart, which allow blood to flow between the left and right blood pools
\cite{Journal:Penny:Lancet2011,Journal:Geva:Lancet2014}. There were 30 2D images for pericardial effusion and 30 2D images for septal defects to provide the positive samples.


For the negative samples, 40 3D CT images from 40 patients without the diseases of interest were used. For each 3D image, six anatomical labels were manually segmented by a radiologist, including the background (BG), right atrium (RA), left atrium (LA), right ventricle (RV), left ventricle (LV), and myocardium (Myo), thus $N = 6$ in Fig. \ref{fig:segnet}. As the four chamber view spans multiple axial slices in a 3D image, multiple slices were extracted from each case totaling 425 2D images. All positive and negative samples were resized to 256$\times$256 pixels.

\begin{table}[t]
\caption{Segmentation results. Average Dice coefficients of different labels from four-fold cross-validations. $n$ is the number of feature channels in the first convolution layer. Format: mean$\pm$std.}
\label{table:dice}
\scriptsize
\centering
\begin{tabularx}{\linewidth}{XXXXXXX}
\toprule
 & BG & RA & LA & RV & LV & Myo \\
\midrule
$n$ = 8 & 96$\pm$0\% & 83$\pm$4\% & 71$\pm$29\% & 82$\pm$6\% & 72$\pm$9\% & 75$\pm$11\% \\
$n$ = 16 & 96$\pm$1\% & 86$\pm$3\% & 88$\pm$2\% & 86$\pm$3\% & 76$\pm$5\% & 83$\pm$3\% \\
$n$ = 32 & 96$\pm$1\% & 85$\pm$4\% & 85$\pm$4\% & 84$\pm$5\% & 76$\pm$5\% & 79$\pm$4\% \\
$n$ = 64 & 96$\pm$1\% & 85$\pm$3\% & 86$\pm$4\% & 86$\pm$2\% & 77$\pm$5\% & 82$\pm$5\% \\
\bottomrule
\end{tabularx}
\end{table}

\section{Experiments and Results on Cardiac CT Data}

\subsection{Segmentation}
\label{sec:exp:seg}
\textbf{Training and validations of the segmentation network on CT data: }Note that only negative cases were used to train the segmentation network. Four-fold cross validations were performed. For each test, images from 30 patients ($\sim$ 300 2D images) were used for training and the rest ($\sim$ 100 2D images) were used for testing. To validate the segmentation accuracy, the Dice coefficients of all six labels between the segmentation and the ground truth were computed. To decide the number of channels for the first convolution layer ($n$), experiments were performed with $n$ = {8, 16, 32, 64}.  We trained the network with a batch size of 10, 100 batches per epoch, and 20 epochs.

\begin{figure}[t]
    \scriptsize
    \centering
    \begin{minipage}[b]{\linewidth}
      \centering
      \begin{minipage}[b]{1.9cm}
      \includegraphics[width=1\linewidth]{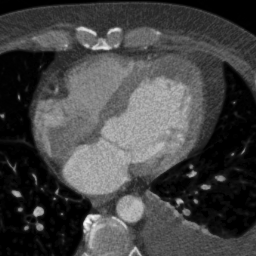}
      \centering{Image}
      \end{minipage}
      \begin{minipage}[b]{1.9cm}
      \includegraphics[width=1\linewidth]{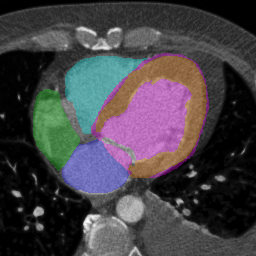}
      \centering{Ground truth}
      \end{minipage}
      \centering
      \begin{minipage}[b]{1.9cm}
      \includegraphics[width=1\linewidth]{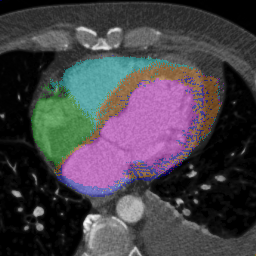}
      \centering{Seg $n$=8}
      \end{minipage}
      \begin{minipage}[b]{1.9cm}
      \includegraphics[width=1\linewidth]{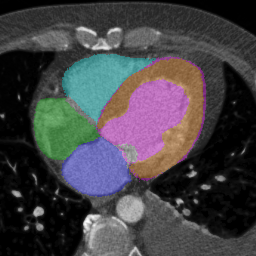}
      \centering{Seg $n$=16}
      \end{minipage}
      \begin{minipage}[b]{1.9cm}
      \includegraphics[width=1\linewidth]{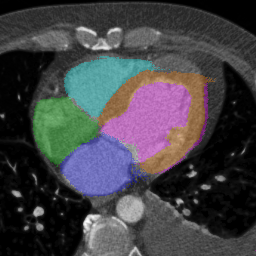}
      \centering{Seg $n$=32}
      \end{minipage}
      \begin{minipage}[b]{1.9cm}
      \includegraphics[width=1\linewidth]{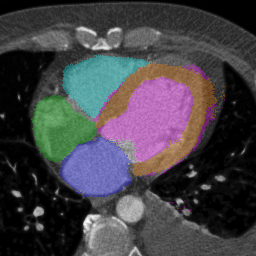}
      \centering{Seg $n$=64}
      \end{minipage} \\
    \end{minipage}
    \\
    \centering{(a) Negative sample.}
    \\
    \vspace{3mm}
    \begin{minipage}[b]{0.49\linewidth}
      \centering
      \begin{minipage}[b]{2cm}
      \includegraphics[width=1\linewidth]{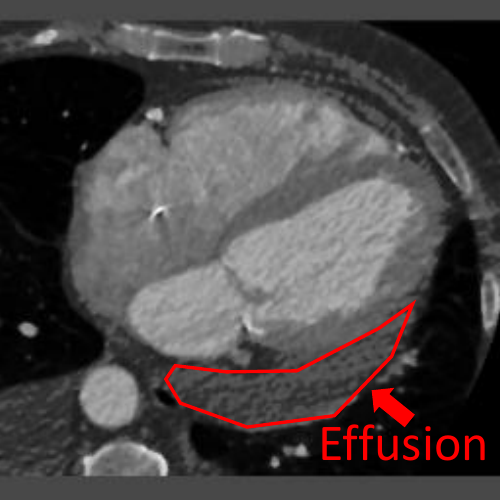}
      \centering{Image}
      \end{minipage}
      \begin{minipage}[b]{2cm}
      \includegraphics[width=1\linewidth]{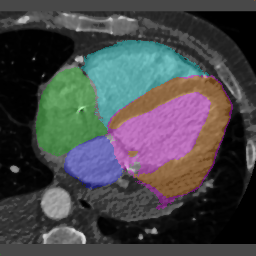}
      \centering{Seg $n$=16}
      \end{minipage} \\
      \centering
      (b) Pericardial effusion.
    \end{minipage}
    \vline
    \begin{minipage}[b]{0.49\linewidth}
      \centering
      \begin{minipage}[b]{2cm}
      \includegraphics[width=1\linewidth]{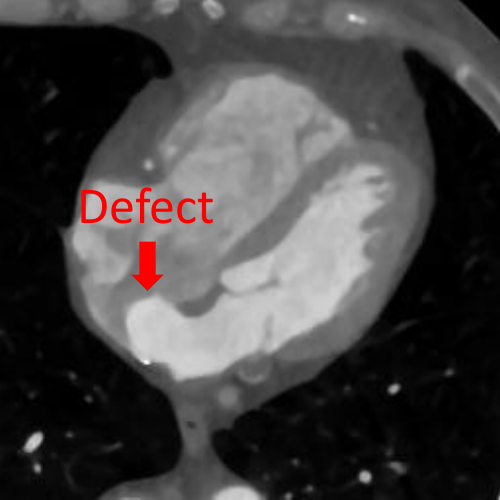}
      \centering{Image}
      \end{minipage}
      \begin{minipage}[b]{2cm}
      \includegraphics[width=1\linewidth]{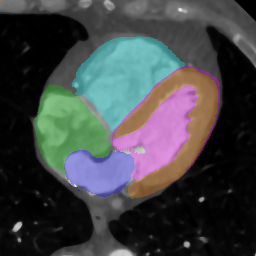}
      \centering{Seg $n$=16}
      \end{minipage} \\
      \centering
      (c) Septal defect.
    \end{minipage}
    \caption{Segmentation on unseen samples. (a) A negative sample. Results with different numbers of feature channels in the first convolution layer ($n$) are shown. (b) and (c) Positive samples of pericardial effusion and septal defect (no ground truth).}
    \label{fig:segnet_seg}
\end{figure}

\textbf{Segmentation results on CT data: }The average results of the cross-validations are shown in Table. \ref{table:dice}. Regardless of the values of $n$, the background was accurately segmented. The Dice coefficients for other labels were lower but still highly accurate. The worst performance happened when $n$ = 8, which had the smallest means and largest standard deviations. For other values of $n$, the standard deviations were not larger than 5\%, thus the performance was very consistent among the validations. Therefore, we chose $n=16$ for the combined architecture as it requires the least parameters. Fig. \ref{fig:segnet_seg} shows the segmentation results. For the unseen negative image (Fig. \ref{fig:segnet_seg}(a)), the segmentation was very similar to the ground truth especially with $n$=16. Consistent with Table. \ref{table:dice}, $n$=8 provided the worst segmentation. For pericardial effusion (Fig. \ref{fig:segnet_seg}(b)), the myocardium segmentation (orange) partially captured the effusion. For septal defect (Fig. \ref{fig:segnet_seg}(c)), the right atrium segmentation (purple) partially captured the defected shape. Therefore, the segmentation network can provide useful semantic and shape information for disease detection, without being trained on them.

\begin{figure}[t]
    \scriptsize
    \centering
    \begin{minipage}[b]{0.32\linewidth}
      \centering
      \centerline{\includegraphics[width=1.1\linewidth]{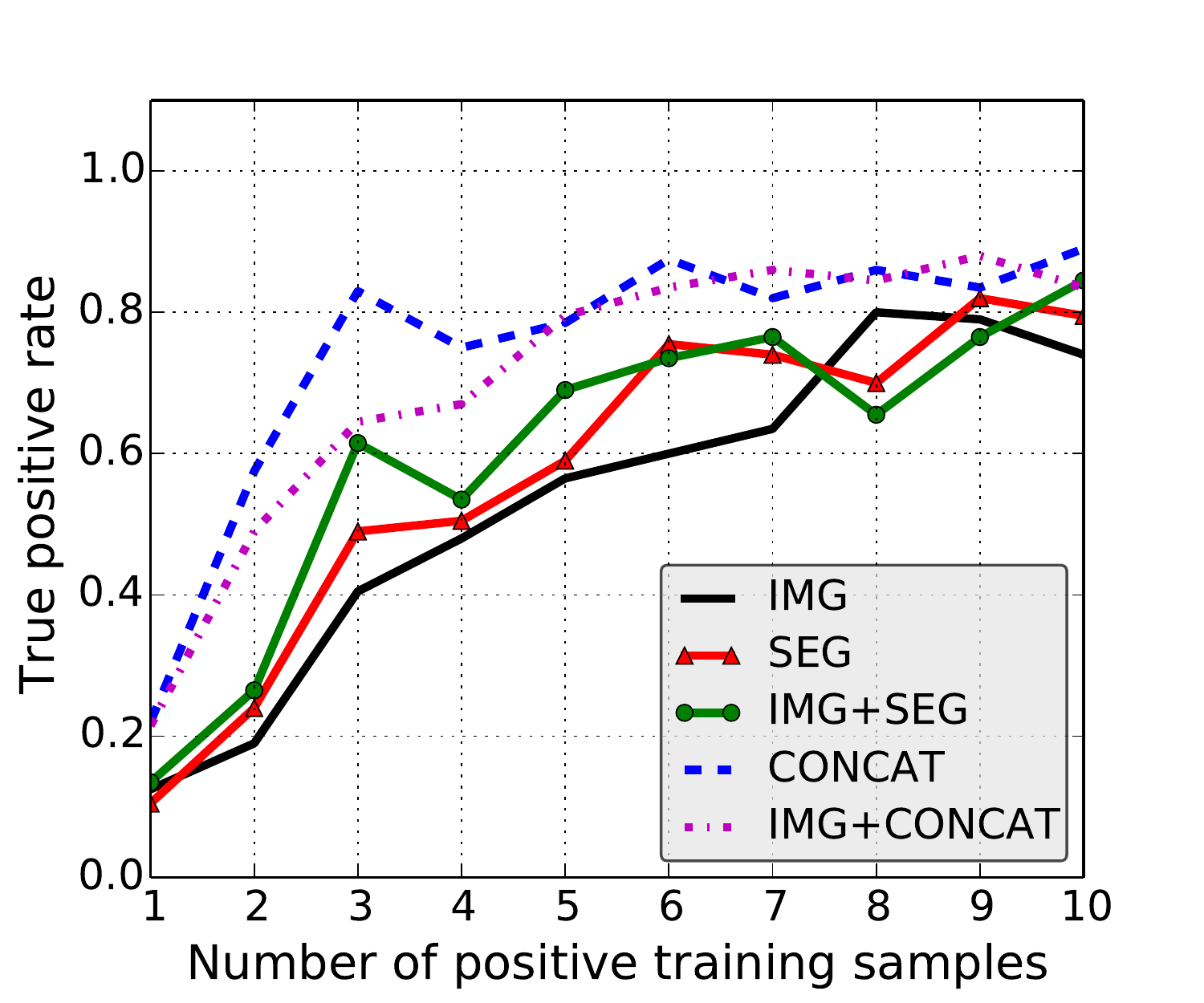}}
    \end{minipage}
    \begin{minipage}[b]{0.32\linewidth}
      \centering
      \centerline{\includegraphics[width=1.1\linewidth]{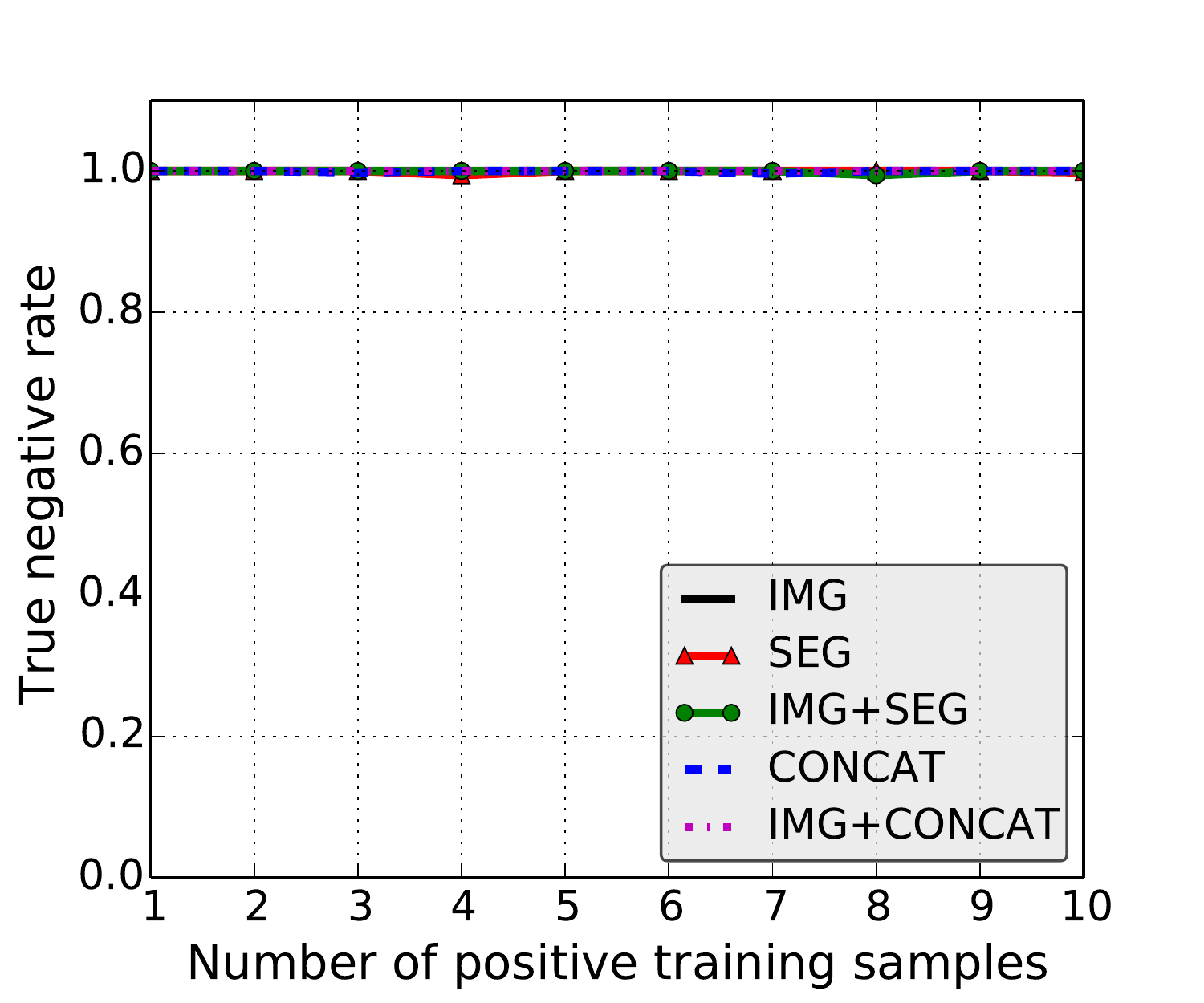}}
    \end{minipage}
    \begin{minipage}[b]{0.32\linewidth}
      \centering
      \centerline{\includegraphics[width=1.1\linewidth]{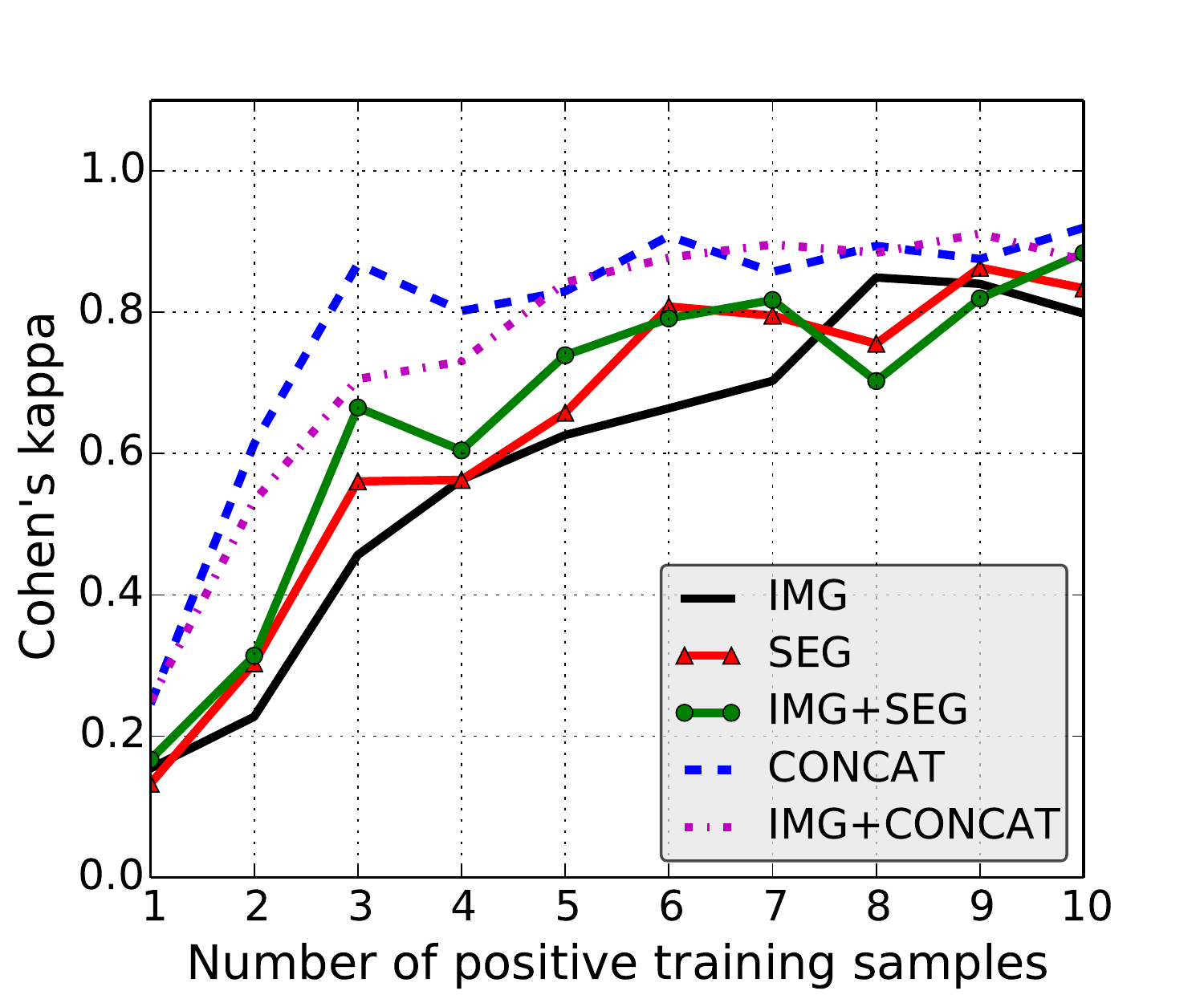}}
    \end{minipage}
    \centering{(a) Pericardial effusion.} \\
    \centering
    \begin{minipage}[b]{0.32\linewidth}
      \centering
      \centerline{\includegraphics[width=1.1\linewidth]{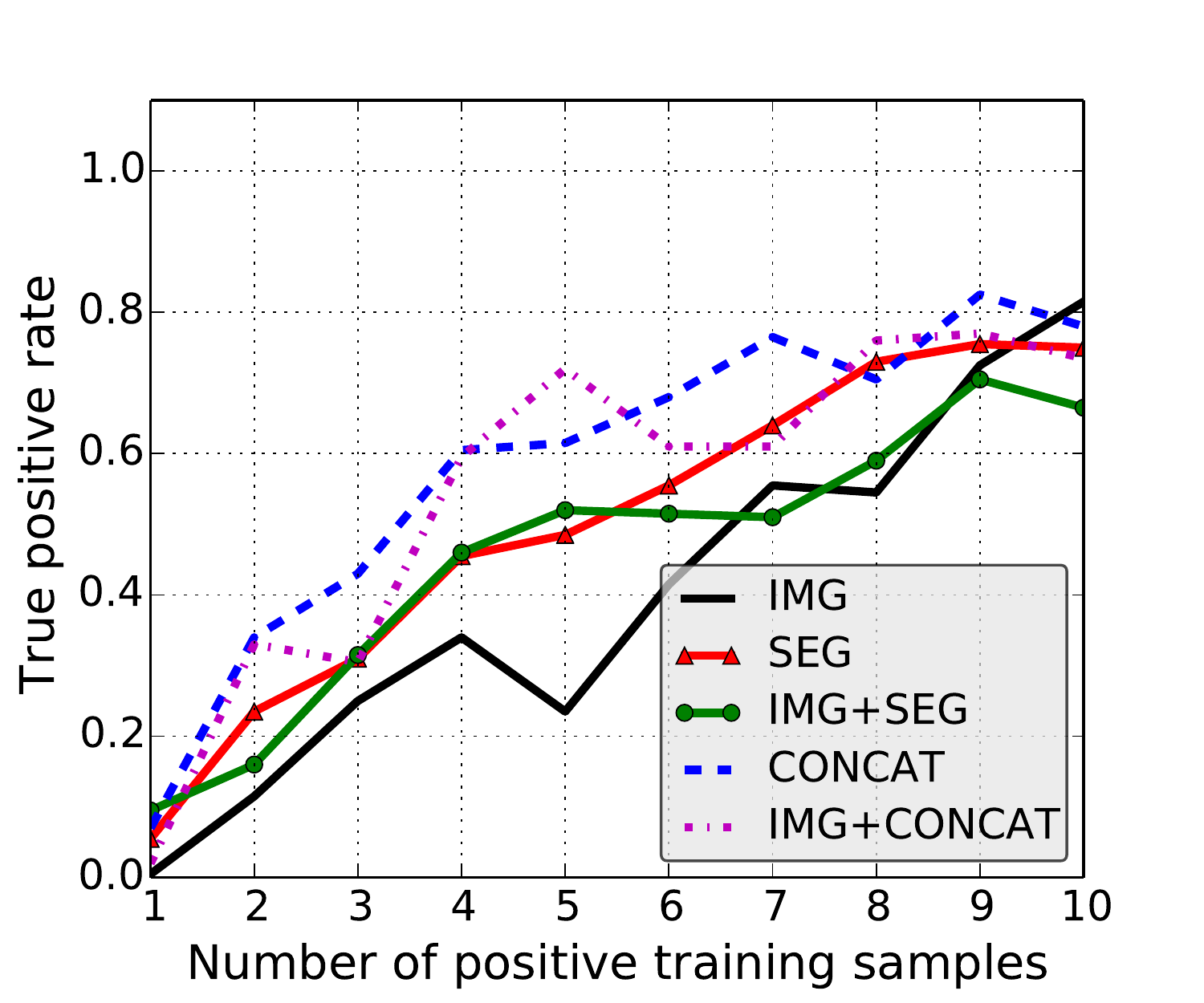}}
    \end{minipage}
    \begin{minipage}[b]{0.32\linewidth}
      \centering
      \centerline{\includegraphics[width=1.1\linewidth]{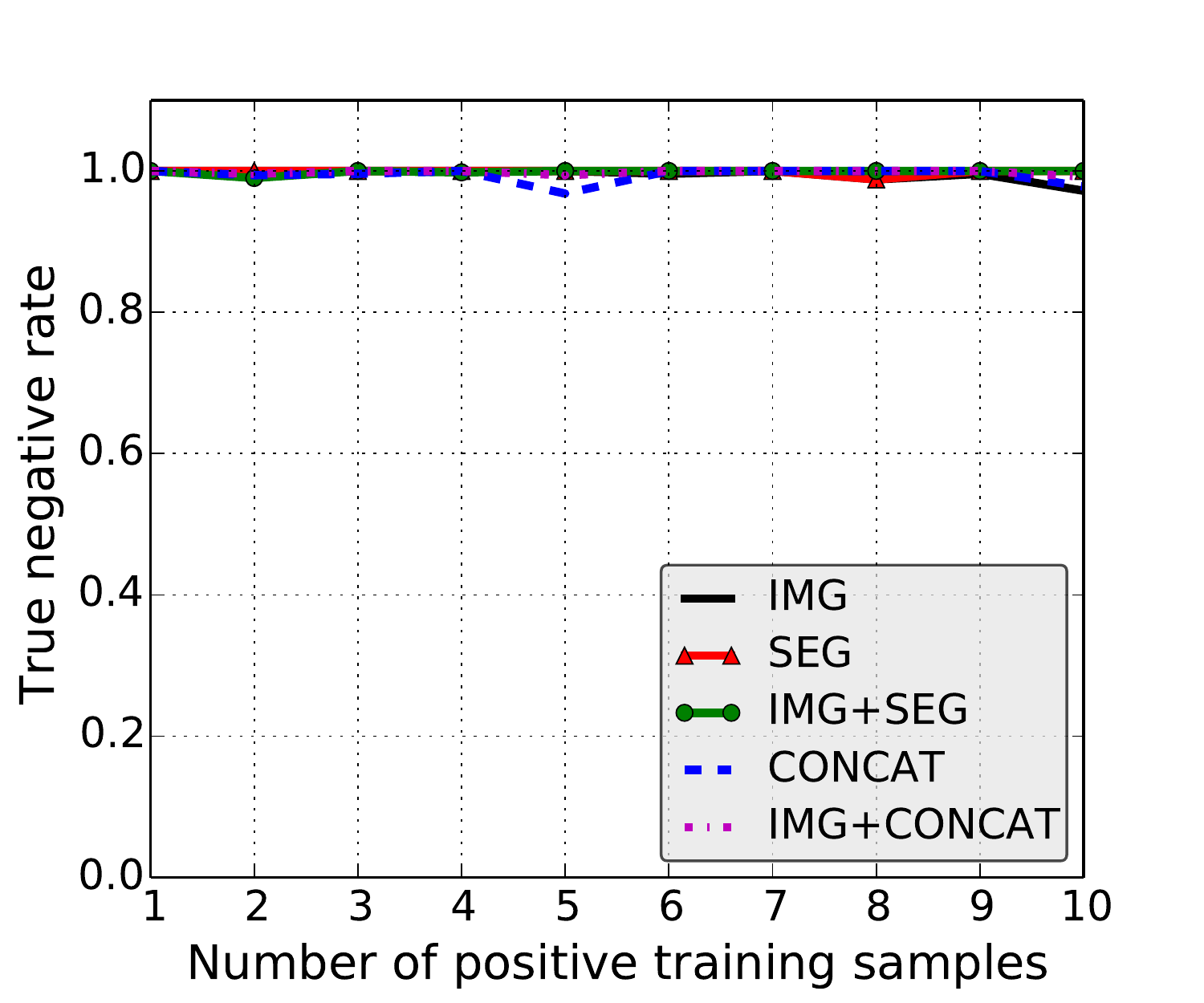}}
    \end{minipage}
    \begin{minipage}[b]{0.32\linewidth}
      \centering
      \centerline{\includegraphics[width=1.1\linewidth]{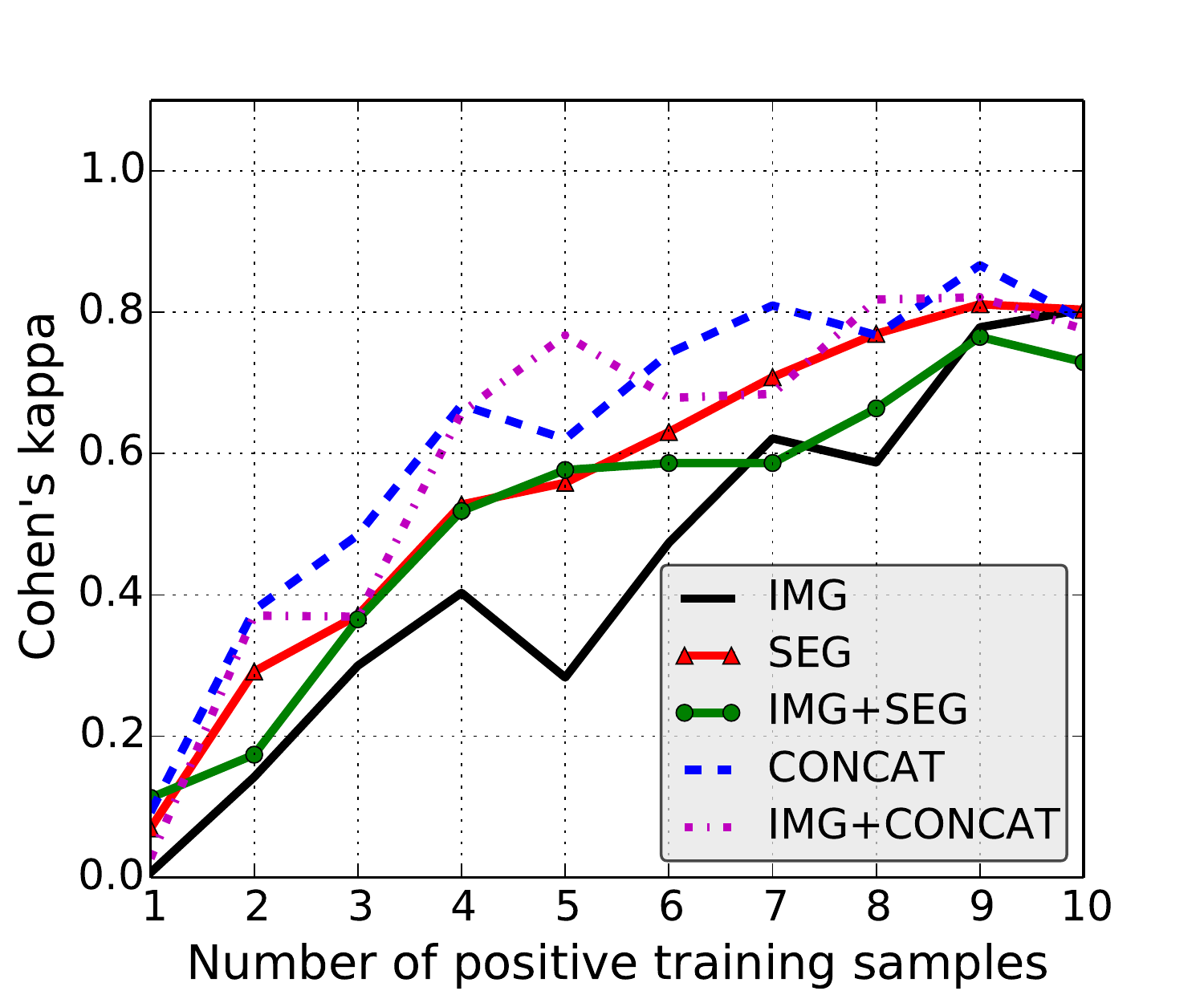}}
    \end{minipage}
    \centering{(b) Septal defects.} \\
    \centering
    \begin{minipage}[b]{0.32\linewidth}
      \centering
      \centerline{True positive rate (recall)}
    \end{minipage}
    \begin{minipage}[b]{0.32\linewidth}
      \centering
      \centerline{True negative rate (specificity)}
    \end{minipage}
    \begin{minipage}[b]{0.32\linewidth}
      \centering
      \centerline{Cohen's kappa}
    \end{minipage}
    \caption{Classification results of different feature combinations with different numbers of positive training samples. The number of negative training samples was 50. The numbers of negative and positive testing samples were 50 and 20, respectively. Each data point represents the average value from 10 experiments.}
    \label{fig:statistics}
\end{figure}

\subsection{Disease Detection}

\textbf{Training and validations of the classification network: }Our goal is to investigate if a segmentation network trained only on negative samples can improve disease detection accuracy with only a few positive training samples. Because of its high accuracy and small number of parameters, a trained segmentation network with $n=16$ was used as the source of features, and the corresponding 100 testing images from 10 patients were used as the negative samples. For each test, the negative samples were randomly divided into two equal sized (50 images) sets for training and testing the classification network. For the 30 positive samples of each disease, they were randomly divided into 10 samples for training and 20 samples for testing. To study the performance with respect to the number of positive samples, we trained 10 models with one to 10 positive samples. Ten repetitions with randomly divided samples were performed for statistical significance, thus 100 tests were performed for each feature combination described in Section \ref{sec:combinations}. The true positive rate (recall), true negative rate (specificity), and Cohen's kappa were used for evaluation. Cohen's kappa provides the inter-rater agreement between the ground truth and detected disease classes. We trained the network with a batch size of 10, 10 batches per epoch, and 20 epochs.

\textbf{Disease detection results on CT data: }Fig. \ref{fig:statistics} shows the classification performances. For both pericardial effusion and septal defect detections, as the number of negative training samples was dominant, the true negative rates were nearly one for all feature combinations. The performances can be distinguished by the true positive rate and the Cohen's kappa. When only the original CT slice was used as an input (IMG), the performance was severely affected by the small number of positive samples. In contrast, when both low and high level features from the segmentation network were included (CONCAT and IMG+CONCAT), the overall performance had a marked increase at the smaller numbers of positive samples compared to IMG. When only the high level segmentation features were included (SEG and IMG+SEG), the performance beats IMG, but not CONCAT. For both SEG and CONCAT, the use of the CT slice itself did not have important effect on the results. This is especially true for CONCAT as the low level features were already included. Although the pericardial effusion detection performed better than the septal defect detection, the differences among feature combinations were consistent. For pericardial effusion, CONCAT with only three positive training samples, a ratio of one to 17 for positive versus negative cases, achieved an average true positive rate of 83\% and Cohen's kappa of 87\%, compared to 41\% and 46\% for IMG. For septal defect, CONCAT trained with seven positive samples yielded an average true positive rate of 77\% and Cohen's kappa of 81\%, compared to 56\% and 62\% for IMG.

\section{Conclusion}

We propose a framework for training disease detection models with very few positive samples. We show that the use of features from a segmentation network results in accurate disease detection and reduces the number of required positive samples to obtain a given level of accuracy. The segmentation network is trained on normal images only, but produces features for both normal and diseased cases. We show notable gains in true positive detection rates using these feature maps on a classification network compared to using only the original images with the same network. Our current data include cases of two cardiac diseases which are each detected on a different classifier. A more general version of this architecture for handling many disease types can be built by adding logic to combine multiple two-class classifiers, or by training a multiclass disease detector.

\bibliographystyle{splncs03}
\bibliography{Ref}

\end{document}